\pdfoutput=1
\documentclass{article}
\PassOptionsToPackage{numbers,square}{natbib}
\usepackage[bottom]{footmisc}
\usepackage{dualverse}
\usepackage[utf8]{inputenc}
\usepackage[T1]{fontenc}
\usepackage{lmodern}
\usepackage{hyperref,url,booktabs,amsfonts,amsmath,amssymb,amsthm,nicefrac,microtype,xcolor,graphicx,multirow}
\usepackage{placeins,needspace}
\usepackage{float}
\newtheorem{proposition}{Proposition}

\newcommand{\fig}[2]{\includegraphics[width=\linewidth]{figures/#1}}
\newcommand{\sys}{PocketRL}
\newcommand{\R}{\mathbb{R}}
\newcommand{\E}{\mathbb{E}}
\newcommand{\fro}[1]{\lVert #1\rVert_F}
\hypersetup{pdftitle={PocketRL: Training a 7B Language Model with Reinforcement Learning on One RTX 5090}}
\title{PocketRL: Training a 7B Language Model with Reinforcement Learning on One RTX~5090}
\SetPaperCode{https://github.com/dualverse-ai/pocketrl}
\SetPaperDate{27-Nov-2026}
\SetCorrespondence{info@dualverse.ai}
\SetPaperLogo{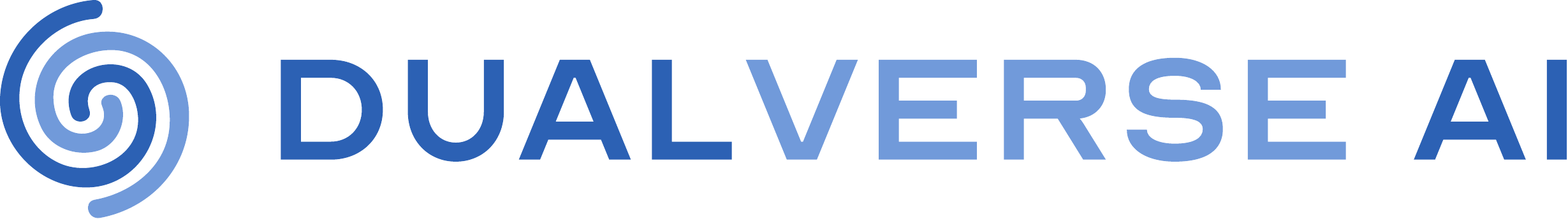}
\SetHeaderTitle{PocketRL: Training a 7B Language Model with Reinforcement Learning on One RTX 5090}
\SetPDFAuthor{Stephen Chung, Wenyu Du}
\SetDualverseAuthors{%
  Stephen Chung\\DualverseAI%
  \and Wenyu Du\\DualverseAI%
}{%
  {\sffamily\bfseries\color{DualverseA}%
    Stephen Chung \quad\quad Wenyu Du%
  }\\[1em]
  \parbox{0.9\linewidth}{\centering
    \normalfont\sffamily\bfseries\small\color{DualverseB}%
    DualverseAI%
  }%
}
\begin{document}\raggedbottom

\begin{abstract}
It is now common to post-train large language models with reinforcement learning with verifiable rewards (RLVR), but its heavy memory use, and hence high cost, is an important bottleneck for RLVR research and a major expense in training frontier models. We introduce \textbf{\sys{}}, a low-cost, memory-efficient RLVR code base that trains a 7B model on a single RTX~5090 and a 27B model on only four, without CPU offloading. \sys{} combines two ideas. First, we project each layer's gradient and optimizer state onto a small basis fitted to the layer's input activations by a QR factorisation, which we call ActGRA, and show that such a basis retains far more of the gradient than the random projections used in prior work. Second, we propose training on 16-bit weights with half an extra byte per weight that carries the rounding error, which prevents small updates from being lost. The same 16-bit weights are used for inference, which saves further memory. Together, these methods reduce memory use for training a 7B model by 84\%. This enables us to train a competitive 7B model in less than 8 hours on a single RTX~5090 for just \$4.71, less than 9\% of the cost of training comparable models. PocketRL also reduces the cost of a 27B training step by more than 77\% compared with similar code bases. We open-source our PocketRL code base.
\end{abstract}
\maketitle
\begingroup
\renewcommand{\thefootnote}{}
\footnotetext[1]{This work extends a human-verified finding made by an \href{https://rstation.ai/}{rStation} agent as part of our \href{https://dualverse.ai/auto_research}{Autonomous Research Program}. See Section~\ref{sec:ai-use} for a discussion of AI use.}
\endgroup
\begin{figure}[b!]
\centering
\includegraphics[width=.91\linewidth]{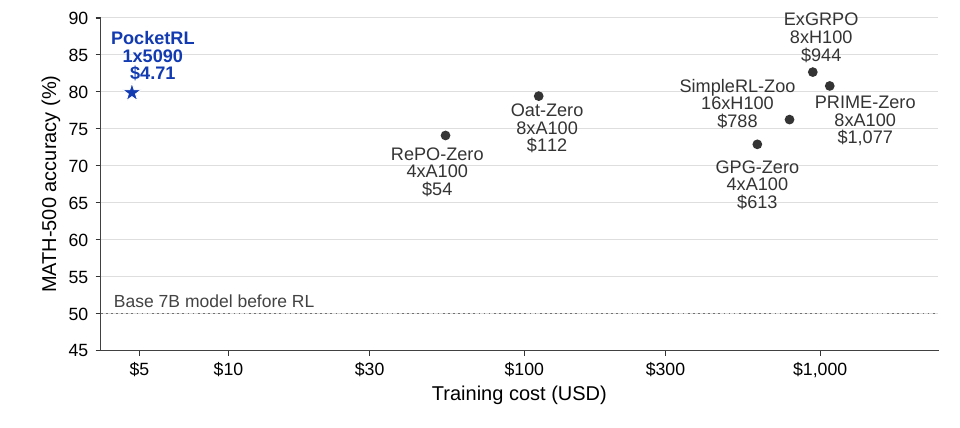}
\caption{\DualverseCaptionTitle{Training cost against accuracy of released 7B models.} MATH-500 accuracy and estimated cost of the training run for \sys{}-7B and six released models trained from the same base model. The base model uses greedy decoding (50.0\%) because its accuracy is lower with sampling (30.2\%). All seven trained models use avg@8 at temperature 1. Table~\ref{tab:main} gives the scores on all six test sets.}
\label{fig:released}
\end{figure}

\section{Introduction}

Reinforcement learning with verifiable rewards (RLVR), in which a model is rewarded when a program verifies its answer, is now a standard stage in the post-training of large language models \citep{r1}\citep{grpo}. It is also memory-intensive: Oat-Zero and SimpleRL-Zoo train 7B models on eight A100s and sixteen H100s, respectively \citep{drgrpo}\citep{simplerl}, and recent work shows that a 27B model can be fitted on eight H100s \citep{logra}. The memory goes mainly to two things: (i) the optimizer state of the dense gradient update, which for Adam is several times the size of the weights, and (ii) the weights themselves, which training typically requires in 32-bit precision so that small updates are not lost. Public code bases also usually keep a separate 16-bit copy of the weights for inference, which speeds up generation but uses more memory. An important question is whether these memory costs can be reduced without losing accuracy or speed, so that RLVR runs on far fewer GPUs or on consumer-grade GPUs.

\begin{figure}[t!]
\centering
\fig{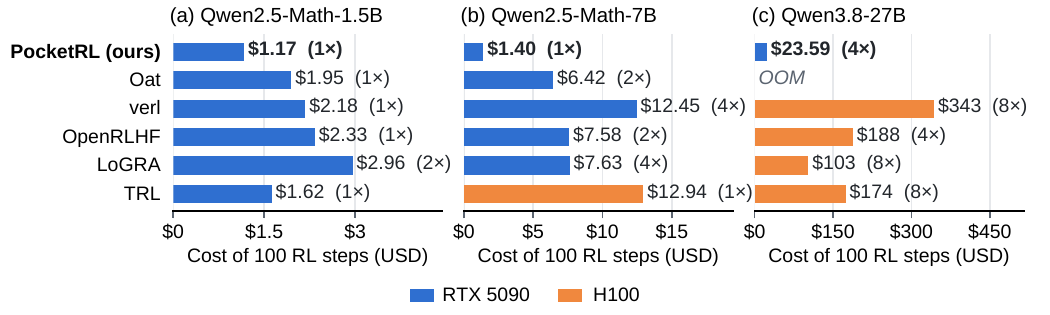}{fig1_cost_a.png}
\caption{\DualverseCaptionTitle{Cost of 100 RL steps in common RLVR code bases.} Each bar is the cheaper of a code base's RTX~5090 and H100 configurations, with GPUs selected according to optimized budget available (GPU count in parentheses; Table~\ref{tab:a1}). Each RL step uses the same workload: 16 prompts with eight sampled answers each, of up to 3,000 tokens.}
\label{fig:cost}
\end{figure}

For the first problem, the optimizer state, LoRA \citep{lora} is a potential mitigation. LoRA trains a small low-rank adapter in place of the weights, which removes most of the optimizer state. However, this confines the weights to a fixed low-rank change. To allow every weight to be trained, a promising direction is gradient compression \citep{galore}\citep{flora}, which stores the gradient and optimizer state in a low-rank projection. LoGRA \citep{logra} is one example, in which the gradient and optimizer state are compressed with random projections, which saves more memory than LoRA at the same rank. However, a random projection retains only a small fraction of each gradient and may reduce accuracy as a result.

To improve the fraction of the gradient retained in gradient compression, we propose ActGRA, in which the projection is fitted to each layer's input activations instead of being random. The gradient of a layer is a weighted sum of the layer's activations, and these activations are concentrated in a few directions, so the gradient often concentrates in the same directions. Therefore, ActGRA finds these activation directions with one QR factorisation and projects the gradient onto them. At rank 128, ActGRA retains 45\% of the gradient, against 7\% for a random projection, and reaches higher accuracy than LoGRA (79.9 $\pm$ 0.6\% against 77.6 $\pm$ 0.4\% on MATH-500 at update 500, 7B model).

For the second problem, the 32-bit training weights, we propose to replace the training weights with 16-bit ones and to use half an extra byte per weight to store each update at higher precision, so that small updates are still retained. Since inference typically uses 16-bit weights, the same 16-bit training weights can be used directly for inference, saving a separate copy of the weights. Together with ActGRA, this lowers the persistent memory cost of training a 7B model by 84\% compared with standard dense training (Table~\ref{tab:budget}).

We combine all of the above into \sys{}, a low-cost RLVR code base. \sys{} trains a 7B model on a single RTX~5090 and a 27B model on only four. \sys{} saves 47\% to 88\% of the cost of a training step compared with the average cost across other code bases at each model size (Figure~\ref{fig:cost}). \sys{}-7B achieves similar performance to other released 7B models for \$4.71, or 8.7\% of the training cost of the next-cheapest model (\$54, Figure~\ref{fig:released}). Our contributions are as follows.
\begin{itemize}
\item \textbf{Open-source repository.} We release \sys{}, a low-cost, efficient RLVR code base that reduces the cost of a training step by 47\% to 88\% and that the research community can build on.
\item \textbf{Significant memory reduction.} Our method reduces memory use for training a 7B model by 84\%, enabling us to train a 7B model on one RTX~5090 and a 27B model on four RTX~5090s.
\item \textbf{Better gradient compression.} We propose ActGRA, a gradient compression method that retains more of the gradient than the random projections used in prior work. We also provide a theoretical analysis of ActGRA.
\item \textbf{Competitive model.} \sys{}-7B achieves accuracy comparable to that of other released 7B models at less than 9\% of the training cost of comparable models.
\end{itemize}

\section{\sys{}}

\begin{figure}[t]
\centering
\fig{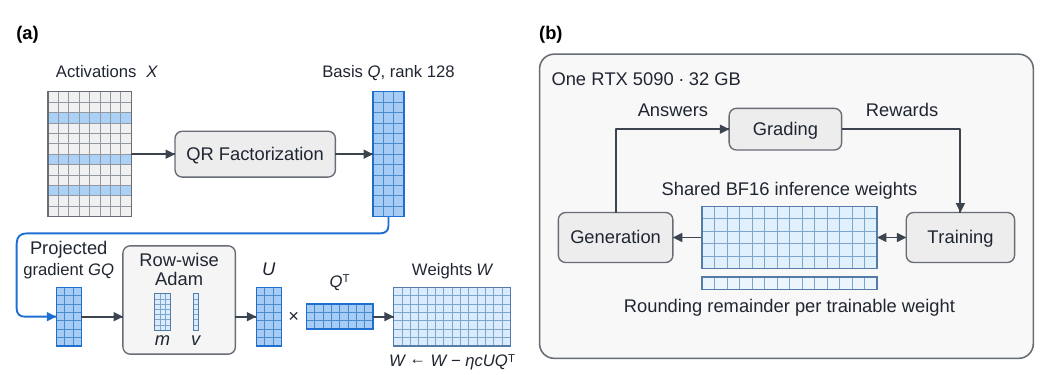}{fig2_method.png}
\caption{\DualverseCaptionTitle{Overview of \sys{}.} (a) ActGRA projects each layer's gradient onto a basis $Q$ fitted to its activations and lifts the update back into the full weight matrix. (b) Generation and training share the 16-bit inference weights. Training keeps a rounding remainder for each weight.}
\label{fig:method}
\end{figure}

\sys{} consists of two key elements (Figure~\ref{fig:method}). The first is ActGRA, which compresses gradients by fitting projection directions to each layer's input activations, reducing the memory needed by the optimiser. The second is to use the same 16-bit weights for training and inference, with half a byte per weight to represent the rounding remainder so small updates are not lost.

\subsection{ActGRA}
\label{sec:update}

\paragraph{Compressing Gradient.} We consider how to compress the gradient during the backward pass. Take a linear layer with weight matrix $W\in\R^{d\times k}$ as an example. Its gradient is
\begin{equation}
G=\sum_{t} g_t x_t^\top \in\R^{d\times k},
\label{eq:grad}
\end{equation}
where the sum runs over all tokens $t$ in the batch, $x_t\in\R^k$ is the layer's input activation at token $t$, and $g_t\in\R^d$ is the gradient of the loss with respect to the layer's output at that token. Standard training with Adam \citep{adam} stores $G$ and two moment buffers of the same size as $W$ for every matrix, leading to high memory usage.

To reduce memory usage, gradient compression stores the gradient in a lower-dimensional space by projecting it along $r\ll k$ directions. Let $Q\in\R^{k\times r}$ be the projection basis whose columns are these $r$ orthonormal directions. The coordinates of the projected gradient in the span of $Q$ are
\begin{equation}
S=GQ=\sum_t g_t\,(Q^\top x_t)^\top\in\R^{d\times r}.
\label{eq:sketch}
\end{equation}
Note that we need not compute $G$ to obtain $S$ during backpropagation, as we simply need to compute the projected activations $Q^\top x_t$, as shown in the second form above. This saves memory because the gradient buffer stores $dr$ entries instead of $dk$.

The remaining question is thus how to choose a good projection basis $Q$. LoGRA \citep{logra} proposes using a random matrix. At each update, it independently samples each entry as $+1/\sqrt{r}$ or $-1/\sqrt{r}$ with equal probability. This random matrix is cheap to obtain, but projecting onto its span discards a large portion of the gradient on average. We thus consider how to choose a better projection basis $Q$ in the following.

\paragraph{Fitting the basis to the activations.} Each row of the weight gradient $G$ is a weighted sum of the layer activations $x_t$ and therefore lies in their span. Because this span generally exceeds $r$ independent directions, we aim to choose $Q$ to capture as much of the activations' squared norm as possible using $r$ orthonormal vectors. To that end, we propose ActGRA, which applies the randomized range finder of \citet{halko}. Specifically, we stack $N$ token activations drawn from sampled sequences of the current batch into the rows of $X_{\mathrm{fit}}\in\R^{N\times k}$, draw an $r\times N$ matrix $\Omega$ of independent standard Gaussian entries, and compute
\begin{equation}
Z=(1/\sqrt{N})\,\Omega X_{\mathrm{fit}},\qquad Z^\top=QR.
\label{eq:qr}
\end{equation}
Each row of $Z$ mixes the sampled layer activations at random, and so directions that account for more of the activations' squared norm tend to appear more strongly in these mixtures. We can then use QR factorization to extract an orthonormal basis $Q$ for the span of these mixtures.

\paragraph{Theoretical Analysis of ActGRA.} To compare projection bases $Q$, we define the retention ratio $\rho(Q)=\fro{GQ}^2/\fro{G}^2$ as the fraction of the squared gradient norm retained by the projection, assuming a nonzero gradient $G$ and orthonormal columns in $Q\in\R^{k\times r}$. A higher retention ratio $\rho(Q)$ means that the projection keeps a larger share of the gradient's squared norm. First we consider a uniformly random basis.

\begin{proposition}[Random basis]\label{prop:random}
Let $G\in\R^{d\times k}$ be any fixed nonzero matrix, and let the orthonormal columns of $Q\in\R^{k\times r}$ span a uniformly distributed $r$-dimensional subspace of $\R^k$. Then the expected retained fraction satisfies
\[
\E[\rho(Q)] = \frac{r}{k}.
\]
\end{proposition}

This shows that when $r$ is much smaller than $k$, a random basis discards most of the gradient's squared norm on average.

Next we consider the activation-fitted basis in ActGRA. Collecting all tokens in the batch into $N$ rows, we stack the layer activations in $X\in\R^{N\times k}$ and the matching output gradients in $\Gamma\in\R^{N\times d}$, so that $G=\Gamma^\top X$. Let $\lVert\Gamma\rVert_{\mathrm{op}}$ denote the spectral norm of $\Gamma$, and let $\sigma_1(X)\ge\sigma_2(X)\ge\dots$ denote the singular values of $X$.

\begin{proposition}[Fitted basis]\label{prop:fitted}
Suppose we compute $Q$ by applying Eq.~\eqref{eq:qr} to all activations in the batch ($X_{\mathrm{fit}}=X$). If the batch is fixed with $G\ne0$ and the integer $1\le q\le r-2$ counts the leading activation directions used for comparison, then the expected retained fraction satisfies
\[
\E[\rho(Q)] \ge 1 - \frac{\lVert\Gamma\rVert_{\mathrm{op}}^2}{\fro{G}^2}\,\frac{r-1}{r-q-1}\sum_{j>q}\sigma_j(X)^2.
\]
If $\operatorname{rank}(X)\le r$, then $\rho(Q)=1$ almost surely.
\end{proposition}

Intuitively, the result shows that the fraction retained by the activation-fitted basis $Q$ depends on how concentrated the activations are. If the activations vary mainly along a few directions, then random Gaussian combinations of the activations mostly lie along those same directions, and the QR factorisation gives a basis $Q$ that tends to capture those directions. ActGRA can therefore retain most of the gradient's squared norm on average, provided the output gradients give little weight to the activation directions that $Q$ misses. Appendix~\ref{app:theory} provides the proofs, and Section~\ref{sec:analysis} reports measurements across layers.

\paragraph{Optimizer and weight update.} \leavevmode{}Given the projected gradient $S=GQ\in\R^{d\times r}$, our Adam variant updates the moments in the compressed space before applying a full-weight step. For the first moment $M$, we replace the exponential moving average of the full gradient with an exponential moving average of $S$. For the second moment $v$, we follow LoGRA's RowAdam \citep{logra} and store one scalar $v_i$ per row to save memory:
\begin{equation}
M\leftarrow\beta_1 M+(1-\beta_1)\,S, \qquad
v_i\leftarrow\beta_2 v_i+(1-\beta_2)\,\tfrac{1}{r}\textstyle\sum_{j=1}^r S_{ij}^2.
\label{eq:moments}
\end{equation}
Using the bias-corrected moments $\hat M$ and $\hat v$, we rescale each row to form $U\in\R^{d\times r}$. We lift $U$ back to $\R^{d\times k}$ via $Q^\top$ and apply the resulting step to $W$:
\begin{equation}
U_{i}=\hat M_{i}\big/\big(\sqrt{\hat v_i}+\epsilon\big), \qquad
W\leftarrow W-\eta\,c\,UQ^\top,
\label{eq:update}
\end{equation}
where $\eta$ is the learning rate and $c=\sqrt{k/r}$ compensates for projecting from $k$ dimensions to $r$. Storing $M$, $v$, and $Q$ requires $dr+d+kr$ scalars per matrix, compared with $2dk$ for standard Adam.

\paragraph{Refitting the basis.} Because layer activations shift during training, we fit a new basis $Q_{\mathrm{new}}$ with Equation~\eqref{eq:qr} every $T$ updates. At a refit, the rows of the first moment $M$ still store coordinates in $Q_{\mathrm{old}}$ and represent the full-space momentum $M Q_{\mathrm{old}}^\top$. Following Flora \citep{flora}, we project this full-space momentum onto $Q_{\mathrm{new}}$ to express it in the new coordinates:
\begin{equation}
M\leftarrow M\,Q_{\mathrm{old}}^\top Q_{\mathrm{new}}.
\label{eq:transport}
\end{equation}
We carry over each $v_i$ unchanged because it summarizes row $i$ with a single scalar and therefore has no basis coordinates to transform. In Appendix~\ref{app:basis-invariance}, we show that any orthonormal basis for the same fitted subspace gives the same weight update, including after a refit.

\subsection{Training with a Quantized Rounding Remainder}
\label{sec:system}

With the small learning rates typically used in RLVR, a small optimizer update can vanish if we use BF16 weights, since rounding may map the updated weight back to its old value \citep{zamirai}. Standard mixed-precision training avoids this loss by keeping an FP32 master copy of each weight \citep{mixedprecision}. In \sys{}, we keep each weight in BF16 and save the rounding difference in a low-bit remainder. Adding this remainder to the next update lets small changes accumulate across steps.

Concretely, for a single weight at step $t$, let $w_t$ be its BF16 value and $\Delta w_t$ its update from Eq.~\eqref{eq:update}. Let $h(w)$ be the gap between adjacent BF16 numbers at the magnitude of $w$. We store the rounding difference as a signed four-bit count $n_t$, where each count represents $h(w_t)/16$. Starting with $n_0=0$, we add the saved difference and the new update in FP32, then round the sum to BF16:
\begin{equation}
u_t = w_t + \frac{n_t}{16}\,h(w_t) + \Delta w_t, \qquad w_{t+1} = \mathrm{round}_{\mathrm{BF16}}(u_t).
\label{eq:remainder-step}
\end{equation}
The difference left over is $u_t-w_{t+1}$. We express this difference in units of $h(w_{t+1})/16$ and stochastically round to an integer ($\mathrm{sround}$) \citep{gupta}:
\[
n_{t+1}=\mathrm{sround}\!\left(\frac{16\,(u_t-w_{t+1})}{h(w_{t+1})}\right),
\]
which is then used in the next update in Eq.~\eqref{eq:remainder-step}. We store the four-bit count $n_t$ alongside the 16-bit weight, for a total of 2.5 bytes per weight.

We combine ActGRA and this quantized rounding remainder in the \sys{} code base. Training and inference share the same weights to save further memory. Together, they lower the persistent memory cost of training a 7B model by 84\%, from 113\,GiB to 18\,GiB (Table~\ref{tab:budget}).

\begin{table}[H]
\centering
\caption{\DualverseCaptionTitle{Memory budget for RLVR on Qwen2.5-Math-7B.}}
\label{tab:budget}
\begin{DVBudgetTable}
& \DVBudgetHead{Standard Training} & \DVBudgetHead{PocketRL} \\
\DVFirstRow{Inference weights (16-bit) & 14.19\,GiB & shared} \\
Training weights (16-bit) & 14.19\,GiB & 14.19\,GiB \\
Master weights (32-bit) & 28.37\,GiB & - \\
Rounding remainder & - & 3.04\,GiB \\
Optimizer state & 56.74\,GiB & 0.61\,GiB \\
\DVBudgetTotalRule
\DVLastRow{Total & 113\,GiB & \textbf{18\,GiB}} \\
\end{DVBudgetTable}
\end{table}

\section{Experiments}
\label{sec:exp}

\subsection{Memory and cost}

We first evaluate the training speed and cost of \sys{} on Qwen2.5-Math-1.5B, Qwen2.5-Math-7B, and Qwen3.8-27B. We compare it with five public RLVR codebases, Oat \citep{oat}, verl \citep{verl}, OpenRLHF \citep{openrlhf}, LoGRA \citep{logra}, and TRL \citep{trl}, on the same RL step with 16 prompts and eight sampled answers per prompt (Table~\ref{tab:a1}). Due to the significant memory reduction, \sys{} is able to train the 7B model on one RTX 5090 and the 27B model on four, without CPU offloading. The other codebases require at least two RTX 5090s at 7B or four H100s at 27B.

At 7B on a single H100, \sys{} completes 100 RL updates in 1.8 hours, 61\% less time than the 4.6 hours required by TRL, the next-fastest codebase. At 27B, \sys{} takes 4.3 hours on four H100s, 9\% more time than LoGRA takes on eight H100s (3.9 hours). Switching to RTX 5090s, which have less memory and a lower price, further reduces the cost of \sys{} by 72\% at 7B and 51\% at 27B. This brings the cost of 100 steps to \$1.40 at 7B and \$23.59 at 27B. These costs are 78\% and 77\% below the next cheapest configurations, Oat at \$6.42 for 7B and LoGRA at \$102.82 for 27B.

\begin{table}[!htb]\centering\small
\caption{\DualverseCaptionTitle{Cost and memory of 100 RL steps.} Each code base uses the smallest number of GPUs, up to eight, that completes an update. Time and USD are for 100 RL steps of the same workload. $^\dagger$CPU offloading.}
\label{tab:a1}
\begin{DVCostTable}
\DVHeaderRow & \DVModelName{Qwen2.5-Math-1.5B} & \DVModelName{Qwen2.5-Math-7B} & \DVModelName{Qwen3.8-27B} \\
\DVModelRules
\DVHeaderRow \DVHead{l}{Code base} & \DVHead{c}{GPUs} & \DVHead{c}{\shortstack{Peak\\GiB/GPU}} & \DVHead{c}{\shortstack{Time\\(h)}} & \DVHead{c}{USD} & \DVHead{c}{GPUs} & \DVHead{c}{\shortstack{Peak\\GiB/GPU}} & \DVHead{c}{\shortstack{Time\\(h)}} & \DVHead{c}{USD} & \DVHead{c}{GPUs} & \DVHead{c}{\shortstack{Peak\\GiB/GPU}} & \DVHead{c}{\shortstack{Time\\(h)}} & \DVHead{c}{USD} \\
\DVGroup{13}{RTX~5090} \\
\textbf{\sys{} (ours)} & 1 & 14.3\phantom{$^\dagger$} & \textbf{1.8} & \textbf{1.17} & 1 & 22.5\phantom{$^\dagger$} & \textbf{2.2} & \textbf{1.40} & 4 & 27.6\phantom{$^\dagger$} & \textbf{9.1} & \textbf{23.59} \\
Oat & 1 & 30.6$^\dagger$ & 3.0 & 1.95 & 2 & 31.3$^\dagger$ & 5.0 & 6.42 & \multicolumn{4}{c}{OOM on 8} \\
verl & 1 & 23.3$^\dagger$ & 3.4 & 2.18 & 4 & 31.4$^\dagger$ & 4.8 & 12.45 & \multicolumn{4}{c}{OOM on 8} \\
OpenRLHF & 1 & 30.1$^\dagger$ & 3.6 & 2.33 & 2 & 31.2$^\dagger$ & 5.9 & 7.58 & \multicolumn{4}{c}{OOM on 8} \\
LoGRA & 2 & 17.6\phantom{$^\dagger$} & 2.3 & 2.96 & 4 & 31.0\phantom{$^\dagger$} & 3.0 & 7.63 & \multicolumn{4}{c}{OOM on 8} \\
TRL & 1 & 24.4\phantom{$^\dagger$} & 2.5 & 1.62 & 2 & 31.4$^\dagger$ & 12.1 & 15.67 & \multicolumn{4}{c}{OOM on 8} \\
\DVGroup{13}{H100} \\
\textbf{\sys{} (ours)} & 1 & 14.4\phantom{$^\dagger$} & \textbf{0.9} & \textbf{2.67} & 1 & 22.5\phantom{$^\dagger$} & \textbf{1.8} & \textbf{4.99} & 4 & 73.9\phantom{$^\dagger$} & 4.3 & \textbf{47.87} \\
Oat & 1 & 51.4\phantom{$^\dagger$} & 1.1 & 3.08 & 1 & 48.2$^\dagger$ & 4.9 & 13.87 & \multicolumn{4}{c}{OOM on 8} \\
verl & 1 & 30.4$^\dagger$ & 2.3 & 6.37 & 1 & 78.3$^\dagger$ & 5.8 & 16.29 & 8 & 67.2$^\dagger$ & 13.1 & 343.22 \\
OpenRLHF & 1 & 34.2$^\dagger$ & 2.1 & 5.99 & 1 & 46.3$^\dagger$ & 10.5 & 29.46 & 4 & 75.6$^\dagger$ & 16.7 & 187.83 \\
LoGRA & 2 & 69.2\phantom{$^\dagger$} & 2.8 & 16.01 & 2 & 69.4\phantom{$^\dagger$} & 9.3 & 52.30 & 8 & 71.9\phantom{$^\dagger$} & \textbf{3.9} & 102.82 \\
\DVLastRow{TRL & 1 & 41.2\phantom{$^\dagger$} & 1.8 & 4.96 & 1 & 78.6\phantom{$^\dagger$} & 4.6 & 12.94 & 8 & 38.8$^\dagger$ & 6.6 & 174.09} \\
\end{DVCostTable}
\end{table}

\subsection{Performance}\label{sec:setup}

We train Qwen2.5-Math-1.5B and Qwen2.5-Math-7B \citep{qwenmath} from their public weights, and Qwen3.8-27B \citep{qwen38} as a scaling test. We train on 8,552 level 3--5 problems from MATH \citep{math} after excluding our held-out problems. We use our \sys{} codebase with Dr.~GRPO \citep{drgrpo} as the base RL algorithm for all main runs, each with three seeds. We compare how gradients are compressed and the respective optimizers, including ActGRA (with the optimizer in Section~\ref{sec:update}), LoGRA (with its proposed RowAdam optimizer), and Dense (uncompressed gradients with standard Adam). For LoGRA, we also use the authors' predicted-KL controller to stabilize training by adjusting the step size according to the predicted KL divergence between the old and updated policies. All 1.5B and 7B runs use one RTX~5090, except Dense at 7B, which requires more memory and uses two H100s. We report avg@8, the fraction of correct answers across eight samples per question at temperature 1, on six mathematics test sets. Appendix~\ref{app:details} gives the hyperparameters and experimental details.

\begin{figure}[t]
\centering
\fig{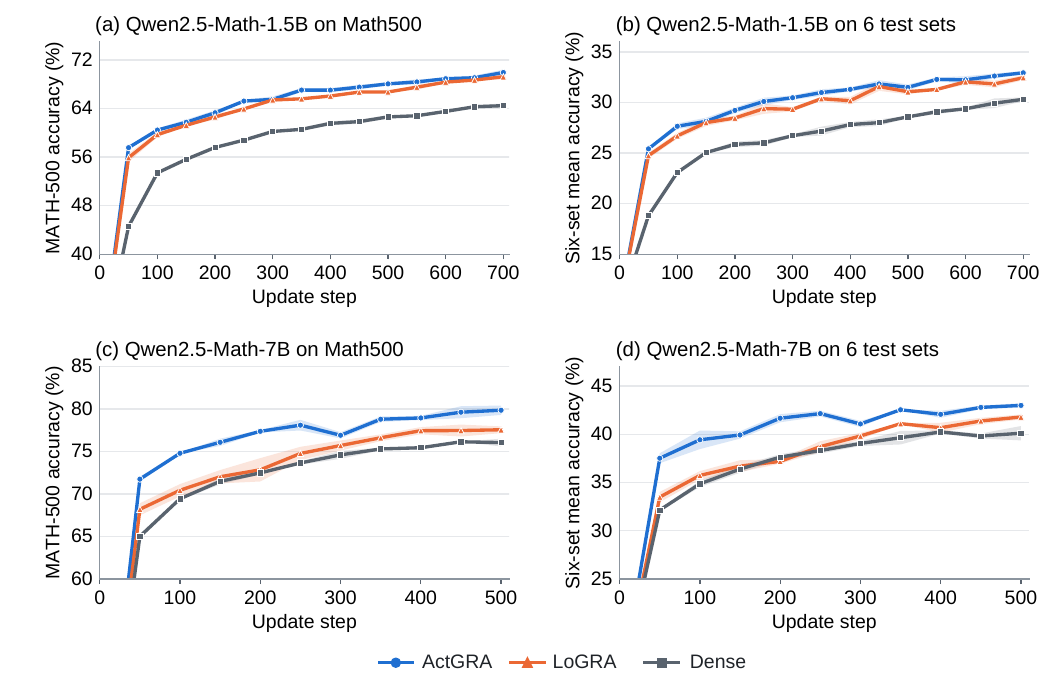}{fig_curves.png}
\caption{\DualverseCaptionTitle{Learning curves.} Avg@8 accuracy on (a, c) MATH-500 and (b, d) the mean of all six mathematics test sets for Qwen2.5-Math-1.5B (top) and Qwen2.5-Math-7B (bottom). Lines show the mean over three seeds; shaded regions show one standard error.}
\label{fig:curves}
\end{figure}

Figure~\ref{fig:curves} shows the learning curves. All three methods improve gradually, with ActGRA performing slightly better than the other two. Table~\ref{tab:main} shows the final-step results. Across the six mathematics test sets at 1.5B, ActGRA finishes at 32.9\% ($\pm$ 0.2), compared with 32.5\% ($\pm$ 0.2) for LoGRA and 30.3\% ($\pm$ 0.2) for Dense. At 7B, ActGRA finishes at 43.0\% ($\pm$ 0.2), compared with 41.8\% ($\pm$ 0.3) for LoGRA and 40.1\% ($\pm$ 0.7) for Dense. Surprisingly, we observe at both 1.5B and 7B that Dense performs worse than ActGRA. We hypothesize that restricting each update to a low-rank subspace reduces fitting to noise in the sampled rewards and helps generalization; similar performance gains from low-rank gradient projection have been observed in previous work \citep{galore}. Altogether, these results suggest that ActGRA compresses gradients with no observed performance loss compared with dense training.

We also compare our trained models with other released RLVR models trained from the same base model. Averaged across the six mathematics test sets, \sys{}-1.5B scores 32.9\%, compared with the released 1.5B mean of 33.2\%, and \sys{}-7B scores 43.0\%, compared with the released 7B mean of 42.1\%. Despite this similar performance, the training cost differs: training the 1.5B model with \sys{} costs \$5.31 and the 7B model \$4.71, while the next cheapest released models cost 18.7$\times$ and 11.5$\times$ as much, demonstrating the cost efficiency of \sys{}.

For the scaling test, we train Qwen3.8-27B for 100 steps. 27B ActGRA improves on the base model by 13.5 percentage points on MATH-500, from 80.85\% to 94.3\% ($\pm$ 0.4), and by 16.9 points across the six mathematics test sets, from 47.87\% to 64.7\% ($\pm$ 0.3) (Table~\ref{tab:main}). This shows that a larger model can also be trained with ActGRA.

\begin{table}[t]
\centering
\caption{\DualverseCaptionTitle{Accuracy on test sets.} We report avg@8 accuracy (\%) as the mean (standard error) over three training seeds at the last step. The Dense, LoGRA and ActGRA rows are trained in \sys{}. We also evaluate released models trained from the same base model. Mem. refers to peak memory. Details on cost estimation can be found in Appendix~\ref{app:cost}.}
\label{tab:main}
\begin{DVResultsTable}
\DVHeaderRow \DVHead{l}{Method} & \DVHead{c}{\shortstack{Math\\500}} & \DVHead{c}{\shortstack{AIME\\2024}} & \DVHead{c}{\shortstack{AIME\\2025}} & \DVHead{c}{\shortstack{AMC\\2023}} & \DVHead{c}{\shortstack{Minerva\\Math}} & \DVHead{c}{\shortstack{Olympiad\\Bench}} & \DVHead{c}{Average} & \DVHead{c}{\shortstack{Est. Cost\\(USD)}} & \DVHead{c}{\shortstack{Mem.\\(GiB)}} & \DVHead{c}{\shortstack{Train\\time (h)}} \\
\DVGroup{11}{Qwen2.5-Math-1.5B} \\
\DVFirstRow{Base model & 21.9 & 2.5 & 1.7 & 16.3 & 5.4 & 12.6 & 10.0 & -- & -- & --} \\
Dense & 64.5 (0.3) & \textbf{13.2 (0.9)} & 5.7 (0.6) & 44.5 (0.4) & 24.8 (0.3) & 29.3 (0.2) & 30.3 (0.2) & 6.53 & 25.6 & 10.1 \\
LoGRA & 69.2 (0.4) & 12.1 (2.1) & 7.6 (0.5) & 45.2 (0.6) & \textbf{27.7 (0.5)} & 32.8 (0.2) & 32.5 (0.2) & 5.66 & 27.4 & 8.8 \\
ActGRA & \textbf{70.0 (0.5)} & 10.8 (0.7) & \textbf{8.5 (0.8)} & \textbf{47.5 (0.7)} & 27.1 (0.3) & \textbf{33.8 (0.4)} & \textbf{32.9 (0.2)} & \textbf{5.31} & \textbf{15.8} & \textbf{8.2} \\
\DVGroup{11}{Released models (1.5B)} \\
\DVFirstRow{Oat-Zero~\citep{drgrpo} & 71.3 & 19.2 & 7.1 & 53.4 & 29.9 & 34.3 & 35.9 & 99 & -- & --} \\
ExGRPO~\citep{exgrpo} & 66.8 & 10.4 & 10.8 & 38.4 & 25.3 & 30.9 & 30.4 & 689 & -- & -- \\
\DVMeanRule
Mean & 69.0 & 14.8 & 9.0 & 45.9 & 27.6 & 32.6 & 33.2 & 394 & -- & -- \\
\DVGroup{11}{Qwen2.5-Math-7B} \\
\DVFirstRow{Base model & 30.2 & 10.8 & 2.5 & 20.0 & 7.2 & 9.5 & 13.4 & -- & -- & --} \\
Dense & 76.1 (0.4) & 22.2 (1.6) & 9.3 (0.9) & 60.6 (1.6) & 33.7 (0.6) & 38.7 (0.2) & 40.1 (0.7) & 46.42 & 74.5 & 7.8 \\
LoGRA & 77.6 (0.4) & \textbf{24.4 (1.0)} & 11.7 (0.4) & 61.7 (0.7) & 34.8 (0.8) & 40.6 (0.6) & 41.8 (0.3) & 7.93 & 31.3 & 12.3 \\
ActGRA & \textbf{79.9 (0.6)} & 23.9 (2.2) & \textbf{12.5 (0.9)} & \textbf{63.2 (1.3)} & \textbf{36.4 (0.5)} & \textbf{42.0 (0.5)} & \textbf{43.0 (0.2)} & \textbf{4.71} & \textbf{22.5} & \textbf{7.3} \\
\DVGroup{11}{Released models (7B)} \\
\DVFirstRow{RePO-Zero~\citep{repo} & 74.1 & 16.7 & 9.6 & 59.7 & 33.0 & 36.9 & 38.3 & 54 & -- & --} \\
Oat-Zero~\citep{drgrpo} & 79.4 & 35.4 & 11.2 & 65.6 & 36.4 & 42.1 & 45.0 & 112 & -- & -- \\
SimpleRL-Zoo~\citep{simplerl} & 76.3 & 24.2 & 11.2 & 61.2 & 34.8 & 38.9 & 41.1 & 788 & -- & -- \\
PRIME-Zero~\citep{prime} & 80.8 & 18.3 & 15.0 & 59.4 & 40.1 & 41.3 & 42.5 & 1077 & -- & -- \\
ExGRPO~\citep{exgrpo} & 82.7 & 25.8 & 15.0 & 63.1 & 38.1 & 46.3 & 45.2 & 944 & -- & -- \\
GPG-Zero~\citep{gpg} & 72.9 & 23.8 & 10.4 & 63.1 & 33.8 & 37.7 & 40.3 & 613 & -- & -- \\
\DVMeanRule
Mean & 77.7 & 24.0 & 12.1 & 62.0 & 36.0 & 40.5 & 42.1 & 598 & -- & -- \\
\DVGroup{11}{Qwen3.8-27B} \\
\DVFirstRow{Base model & 80.8 & 25.8 & 25.4 & 66.6 & 41.2 & 47.4 & 47.9 & -- & -- & --} \\
\DVLastRow{ActGRA & 94.3 (0.4) & 51.1 (1.9) & 44.3 (0.4) & 86.4 (1.9) & 49.8 (0.7) & 62.5 (0.4) & 64.7 (0.3) & 23.53 & 31.1 & 9.1} \\
\end{DVResultsTable}
\end{table}

\subsection{Analysis}\label{sec:analysis}

\begin{figure}[t]
\centering
\includegraphics[width=\linewidth]{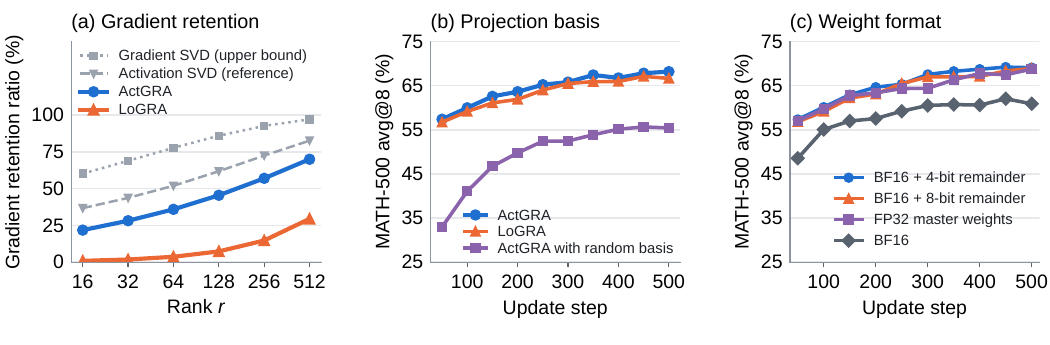}
\caption{\DualverseCaptionTitle{Analysis and ablation studies of ActGRA and the quantized rounding remainder on Qwen2.5-Math-1.5B.} (a) Gradient retention by different gradient compression methods. (b) MATH-500 avg@8 accuracy during training with different gradient compression methods. (c) MATH-500 avg@8 accuracy during training with different weight formats.}
\label{fig:gradshare}
\end{figure}

\DualverseParagraphTitle{Gradient retention.} We then analyze how much of the squared gradient norm ActGRA retains compared with the random projection used in LoGRA, on the 1.5B model (Figure~\ref{fig:gradshare}a). At rank 128, LoGRA's random projection retains 7.3\%, close to the 7.35\% predicted by Proposition~\ref{prop:random}. In contrast, ActGRA retains 45.5\%, consistent with our theoretical analysis that it retains a larger portion of the gradient. As expected, the gradient retention ratio increases with rank $r$ for both methods, though LoGRA requires rank 512 to achieve a similar retention ratio to ActGRA at rank 32.

To isolate the effect of the projection basis on downstream performance, we perform additional training by replacing only the basis $Q$ in ActGRA with the random projection used in LoGRA, without adding LoGRA's predicted-KL controller. Figure~\ref{fig:gradshare}b shows the learning curves. At update 500, MATH-500 avg@8 accuracy falls from 68.2\% to 55.4\% on the 1.5B model. This indicates that ActGRA's fitted basis improves downstream performance over a random basis without requiring an extra predicted-KL controller to stabilize learning.

\Needspace{7\baselineskip}
\DualverseParagraphTitle{Quantized Rounding Remainder.} To evaluate BF16 training weights with the quantized rounding remainder, we compare four matched runs using BF16 with an eight-bit or a four-bit remainder, plain BF16, and FP32 master weights (Figure~\ref{fig:gradshare}c). Removing the four-bit remainder reduces MATH-500 avg@8 accuracy from 69.05\% to 60.9\% at step 500, showing its importance for accumulating small updates. BF16 weights with a four-bit remainder, BF16 weights with an eight-bit remainder, and FP32 master weights all perform similarly, suggesting that a four-bit remainder is sufficient to compensate for the lost precision.

\Needspace{8\baselineskip}
\FloatBarrier
\section{Related work}

\paragraph{RLVR and existing repositories.} GRPO \citep{grpo} and its use in DeepSeek-R1 \citep{r1} made RLVR a common recipe for reasoning models. Dr.~GRPO revises the objective \citep{drgrpo}, SimpleRL-Zoo studies reward design and problem difficulty \citep{simplerl}, and DAPO combines changes to clipping, sampling and reward shaping \citep{dapo}. Thinker trains models to answer in stages, with rewards for the outcomes of each stage \citep{thinker}. We use Dr.~GRPO as the main algorithm while investigating different gradient compression methods. We compare \sys{} with public RLVR implementations in Oat \citep{oat}, verl \citep{verl}, OpenRLHF \citep{openrlhf}, TRL \citep{trl}, and LoGRA \citep{logra}.

\paragraph{Memory-efficient training.} LoRA \citep{lora} and QLoRA \citep{qlora} train a low-rank accumulated update over frozen weights. Gradient compression instead projects each gradient into a low-rank space: GaLore \citep{galore} uses singular vectors of the gradient, and Flora \citep{flora} uses resampled random directions, carrying momentum across. LoGRA \citep{logra}, our closest work, uses random projections in RLVR to compress gradients and optimizer state. ActGRA constructs its projection basis from random combinations of the layer's activations using QR factorization, requiring neither a singular value decomposition nor the full gradient while retaining a large portion of the gradient. Prior work~\citep{compact}\citep{eva}\citep{prac} also uses input activations for memory-efficient training in other settings but differs from our approach, which applies QR factorization to random linear combinations of the activations.

Training with low-precision weights has been explored in prior work~\citep{zamirai}\citep{gupta}\citep{flashoptim}\citep{bonin}. Prior work~\citep{zamirai}\citep{gupta} applies stochastic rounding directly to the weights, while we apply it to the quantized rounding remainder. FlashOptim~\citep{flashoptim} also carries the rounding error across updates during pretraining, but it rounds the correction to the nearest integer and stores it in 8 or 16 bits. Bonin~\citep{bonin} uses the rounding error to train binary weights.

\section{Conclusion}

We presented \sys{}, a low-cost, memory-efficient codebase for RLVR. It combines ActGRA, a novel input-fitted projection method for gradient compression, and a quantized rounding remainder. These significantly reduce the memory usage of RLVR, allowing large models to fit on a few RTX 5090s without CPU offloading. This allows training a 7B model at less than 9\% of the training cost of comparable models and cuts the cost of each training step for a 27B model by more than 77\%.

\FloatBarrier
\section{AI Use}
\label{sec:ai-use}
This work builds on a finding made by an autonomous agent run by the authors on rStation, a multi-agent research platform. Given the goal of training a 7B model on a single RTX~5090 with competitive performance, the agent found the methods introduced in this paper, including ActGRA and the quantized rounding remainder, and built the original code base. The agent published its findings at \url{https://rstation.ai/problems/math_rlvr/papers/40/v1}.

The authors considered this finding potentially useful to the broader research community and therefore reproduced and extended the experiments outside rStation, using more models and more seeds. The authors also revised the manuscript to make the findings accessible to researchers and restructured the agent's code base into the final PocketRL code base.

\bibliographystyle{unsrtnat}
\bibliography{refs}

\appendix

\section{Theoretical Analysis}
\label{app:theory}

Throughout this appendix, let $N$ denote the total number of tokens in a batch. Stacking the layer activations row by row into $X\in\R^{N\times k}$ and the corresponding output gradients into matching rows of $\Gamma\in\R^{N\times d}$ gives the layer weight gradient $G=\Gamma^\top X\in\R^{d\times k}$.

\subsection{Bounds on the Retention Ratio}

We first derive the expected retention ratio $\rho(Q)$ for a uniformly random basis. We then bound the fraction retained by ActGRA's activation-fitted basis $Q$.

\begin{proof}[Proof of Proposition~\ref{prop:random}]
Let $P=QQ^\top\in\R^{k\times k}$ be the orthogonal projector onto the column space of $Q$. By the orthogonal invariance of the uniform distribution over $r$-dimensional subspaces of $\R^k$, the expectation $\E[P]$ commutes with every orthogonal transformation on $\R^k$ and must therefore be a scalar multiple of $I_k$. Because $\mathrm{tr}(P)=r$ identically, taking traces gives $\E[P]=(r/k)I_k$. Writing $\fro{GQ}^2=\mathrm{tr}(GQQ^\top G^\top)=\mathrm{tr}(GPG^\top)$ and applying linearity of expectation and trace yields $\E[\fro{GQ}^2]=\mathrm{tr}(G\,\E[P]\,G^\top)=(r/k)\fro{G}^2$. Dividing both sides by $\fro{G}^2$ completes the proof.

The same $r/k$ retention ratio holds in expectation for orthogonal projection onto the span of LoGRA's random sign matrix, conditional on full rank. Sign flips and coordinate permutations preserve its distribution, so the expected projector is again $(r/k)I_k$.
\end{proof}

\begin{proof}[Proof of Proposition~\ref{prop:fitted}]
Let $P=QQ^\top$. Because $P$ and $I_k-P$ are complementary orthogonal projectors and $Q$ has orthonormal columns, Frobenius orthogonality gives $\fro{G}^2=\fro{GP}^2+\fro{G(I_k-P)}^2=\fro{GQ}^2+\fro{G(I_k-P)}^2$. Writing $G(I_k-P)=\Gamma^\top X(I_k-P)$ and applying the submultiplicative inequality $\fro{AB}\le\lVert A\rVert_{\mathrm{op}}\fro{B}$, we obtain
\[
\fro{G(I_k-P)}^2 \le \lVert\Gamma^\top\rVert_{\mathrm{op}}^2\,\fro{X(I_k-P)}^2 = \lVert\Gamma\rVert_{\mathrm{op}}^2\,\fro{X-XQQ^\top}^2.
\]
Substituting this inequality into $\fro{GQ}^2=\fro{G}^2-\fro{G(I_k-P)}^2$ and dividing by $\fro{G}^2$ yields the deterministic lower bound
\begin{equation}
\rho(Q) \ge 1 - \frac{\lVert\Gamma\rVert_{\mathrm{op}}^2\,\fro{X-XQQ^\top}^2}{\fro{G}^2}.
\label{eq:det-retention}
\end{equation}

Next, set $A=X^\top\in\R^{k\times N}$. With $X_{\mathrm{fit}}=X$ in Eq.~\eqref{eq:qr}, let $Y=\sqrt{N}\,Z^\top=A\Omega^\top$, where $\Omega\in\R^{r\times N}$ has independent $\mathcal{N}(0,1)$ entries. Whenever $\operatorname{rank}(Y)<r$, we extend an orthonormal basis of $\operatorname{range}(Y)$ to $r$ orthonormal columns in $Q$, so $P=QQ^\top$ always projects onto a subspace containing $\operatorname{range}(Y)$. Let $s=\operatorname{rank}(A)$. If $s\le r$, let $V_s\in\R^{N\times s}$ have orthonormal columns spanning the right singular subspace of $A$. Then $V_s^\top\Omega^\top\in\R^{s\times r}$ has independent $\mathcal{N}(0,1)$ entries and therefore has full row rank $s$ almost surely. Consequently, $\operatorname{range}(Y)=\operatorname{range}(A)$ almost surely, which implies $(I_k-P)A=0$, $\fro{X-XP}^2=0$, and hence $\rho(Q)=1$ almost surely.

To prove the expectation bound, note that the case $\operatorname{rank}(X)\le r$ follows immediately from exact recovery, so we assume $\operatorname{rank}(A)>r>q$. Let $A=U\Sigma V^\top$ be a full singular value decomposition with right singular basis $V\in\R^{N\times N}$, and let $\Sigma_2$ denote the trailing block of $\Sigma$ after the first $q$ singular values. Partition $V^\top\Omega^\top$ at row $q$ into $\Omega_1\in\R^{q\times r}$ and $\Omega_2\in\R^{(N-q)\times r}$. By the orthogonal invariance of the standard Gaussian distribution, $\Omega_1$ and $\Omega_2$ are independent standard Gaussian matrices, and $\Omega_1$ has full row rank $q$ almost surely because $r>q$. Applying the structural bound of Theorem~9.1 of \citet{halko} directly to the squared Frobenius norm gives
\[
\fro{X-XP}^2 = \fro{(I_k-P)A}^2 \le \fro{\Sigma_2}^2 + \fro{\Sigma_2\Omega_2\Omega_1^\dagger}^2.
\]
Conditioning on $\Omega_1$, the entries of $\Omega_2$ remain independent standard normal random variables, so
\[
\E\bigl[\fro{\Sigma_2\Omega_2\Omega_1^\dagger}^2 \mid \Omega_1\bigr] = \fro{\Sigma_2}^2\,\fro{\Omega_1^\dagger}^2.
\]
Moreover, $\fro{\Omega_1^\dagger}^2=\mathrm{tr}\bigl((\Omega_1\Omega_1^\top)^{-1}\bigr)$, where $\Omega_1\Omega_1^\top\sim\mathcal{W}_q(I_q,r)$ with $r-q-1\ge 1$. By the first moment of the inverse Wishart distribution, $\E_{\Omega_1}[\fro{\Omega_1^\dagger}^2]=q/(r-q-1)$. Combining these expectations with $1+q/(r-q-1)=(r-1)/(r-q-1)$ and $\fro{\Sigma_2}^2=\sum_{j>q}\sigma_j(X)^2$ yields
\[
\E_\Omega\bigl[\fro{X-XP}^2\bigr] \le \frac{r-1}{r-q-1}\sum_{j>q}\sigma_j(X)^2.
\]
Taking the expectation over $\Omega$ in Eq.~\eqref{eq:det-retention} establishes the claimed inequality.
\end{proof}

\subsection{Conditional Expectations of Squared Gradient Norms}

The next result relates the retention ratio $\rho(Q)$ from Section~\ref{sec:update} to the fraction of the activations' squared norm captured by $Q$.

\begin{proposition}[Gradient retention under isotropic output gradients]
\label{prop:share}
Suppose $X\ne 0$ and the output gradients have an isotropic conditional second moment across tokens, $\E[\Gamma\Gamma^\top\mid X,Q]=\sigma^2 I_N$ with $\sigma^2>0$. Then
\[
\frac{\E\bigl[\fro{G}^2\rho(Q)\mid X,Q\bigr]}{\E\bigl[\fro{G}^2\mid X,Q\bigr]} = \frac{\fro{XQ}^2}{\fro{X}^2}.
\]
\end{proposition}

Intuitively, the weight gradient combines the layer activations using the output gradients. Under the stated assumption, the output gradients have equal average squared norm across tokens, and contributions from different tokens cancel in the cross terms on average. Both the expected squared norm of the projected gradient and that of the full gradient are then proportional to the corresponding squared norms of the activations. This explains why ActGRA fits $Q$ to the layer activations: if $Q$ captures 80\% of their squared norm, it retains 80\% of the expected squared gradient norm.

\begin{proof}[Proof of Proposition~\ref{prop:share}]
By definition, $\fro{G}^2\rho(Q)=\fro{GQ}^2$, taking the product to be zero when $G=0$.
Conditioning on $X$ and $Q$, linearity of expectation and trace together with the isotropic second-moment assumption give $\E[\fro{GQ}^2\mid X,Q]=\E[\fro{\Gamma^\top XQ}^2\mid X,Q]=\E[\mathrm{tr}(Q^\top X^\top\Gamma\Gamma^\top XQ)\mid X,Q]=\sigma^2\fro{XQ}^2$. By the same calculation without $Q$, $\E[\fro{G}^2\mid X,Q]=\E[\fro{\Gamma^\top X}^2\mid X,Q]=\sigma^2\fro{X}^2$. Since $X\ne 0$ and $\sigma^2>0$, the denominator is strictly positive, and dividing the two expectations yields the claimed ratio.
\end{proof}

\subsection{Invariance to a Change of Basis}\label{app:basis-invariance}

The optimizer in Section~\ref{sec:update} stores its first moment in the coordinates of the fitted basis. We show here that the ActGRA weight update is invariant to the choice of orthonormal basis within the fitted subspace: two such bases can give different coordinates for the first moment, yet both produce exactly the same weight update. This also holds after a refit: for a given sequence of fitted subspaces, the choice of coordinates within each subspace leaves the weight updates unchanged.

To verify this, let $O\in\R^{r\times r}$ be any orthogonal matrix and replace $Q$ by $QO$. Under this basis change, the projected gradient $S$ transforms to $SO$ and the first moment $M$ transforms to $MO$. Because right-multiplication by an orthogonal matrix leaves Euclidean row norms invariant, the squared row norms of $S$ and the row-wise second moments $v_i$ are unchanged. Consequently, the rescaled update $U$ transforms to $UO$, and the resulting weight update $UO(QO)^\top=UOO^\top Q^\top=UQ^\top$ is invariant. When we refit the basis, the first-moment update in Eq.~\eqref{eq:transport} sends $MO_{\mathrm{old}}$ to $(MO_{\mathrm{old}})(Q_{\mathrm{old}}O_{\mathrm{old}})^\top(Q_{\mathrm{new}}O_{\mathrm{new}})=(MQ_{\mathrm{old}}^\top Q_{\mathrm{new}})O_{\mathrm{new}}$. Hence, even after a refit, any orthonormal basis for the same fitted subspace gives the same weight update.

\section{Pseudocode}

\begin{figure}[H]
\begin{dvalgorithm}{alg}{ActGRA update for a linear layer $W\in\R^{d\times k}$}
\DVAlgLine & \textbf{Init:} BF16 $W\in\R^{d\times k}$, $M\leftarrow 0\in\R^{d\times r}$, $v\leftarrow 0\in\R^{d}$, 4-bit remainders $n\leftarrow 0$ & \\
\DVAlgLine & \textbf{for} step $s=1,2,\dots$ \textbf{do} & \\
\DVAlgLine & \AlgInd1\textbf{if} $s\equiv 1 \pmod T$ \textbf{then} & \\
\DVAlgLine & \AlgInd2$Z\leftarrow (1/\sqrt{N})\,\Omega X_{\mathrm{fit}}$;\; $Q_{\mathrm{new}}\leftarrow\mathrm{QR}(Z^\top)$ & Eq.~\eqref{eq:qr} \\
\DVAlgLine & \AlgInd2$M\leftarrow MQ^\top Q_{\mathrm{new}}$ (if $s>1$);\; $Q\leftarrow Q_{\mathrm{new}}$ & Eq.~\eqref{eq:transport} \\
\DVAlgLine & \AlgInd1$S\leftarrow \sum_t g_t(Q^\top x_t)^\top$ & Eq.~\eqref{eq:sketch} \\
\DVAlgLine & \AlgInd1 Update $M$ and $v$ from $S$ & Eq.~\eqref{eq:moments} \\
\DVAlgLine & \AlgInd1$U_i\leftarrow \hat M_i/(\sqrt{\hat v_i}+\epsilon)$;\; $\Delta W\leftarrow -\eta\,c\,UQ^\top$ & Eq.~\eqref{eq:update} \\
\DVAlgLine & \AlgInd1 Update BF16 $W$ and 4-bit remainders $n$ with $\Delta W$ & Eq.~\eqref{eq:remainder-step} \\
\DVAlgLine & \textbf{end for} &
\end{dvalgorithm}
\end{figure}

\section{Experiment Details}
\label{app:details}

\subsection{Training Setup and Implementation}

The hyperparameters and detailed training settings for the main runs can be found in Table~\ref{tab:main-settings}.

\begin{table}[H]
\centering\small
\caption{\DualverseCaptionTitle{Training settings for the main runs.}}
\label{tab:main-settings}
\begin{DVSettingsTable}
\DVHeaderRow \DVHead{l}{Setting} & \DVHead{l}{Value} \\
\DVFirstRow{RL objective & Dr.\ GRPO} \\
Updates (1.5B / 7B / 27B) & 700 / 500 / 100 \\
Training seeds & 42, 7, 101 \\
Prompts per rollout / responses per prompt & 16 / 8 \\
Learning rate & $10^{-6}$, constant \\
Gradient-norm clipping & Enabled \\
Context window / maximum response length & 4,096 / 3,000 tokens \\
Sampling temperature / top-$p$ & 1 / 1 \\ \midrule
ActGRA and LoGRA projection rank & 128 \\
ActGRA target matrices & Attention and MLP weights \\
ActGRA basis refit & Every 8 updates on sampled batch activations \\
ActGRA optimizer & $\beta_1=0.9$, $\beta_2=0.95$, $\epsilon=10^{-8}$ \\
ActGRA perturbation to $Z$ before QR & Random signs scaled by $10^{-3}$ times the mean row norm \\
LoGRA optimizer and controller & RowAdam and predicted-KL control \\
\DVLastRow{Dense optimizer & Adam with FP32 master weights} \\
\end{DVSettingsTable}
\end{table}

\paragraph{One-byte rounding remainder at 1.5B.}
Because the 1.5B model fits on a single RTX~5090, our main ActGRA runs at this size allocate an 8-bit (one-byte) rounding remainder per weight instead of the 4-bit remainder in Section~\ref{sec:system}. In Eq.~\eqref{eq:remainder-step}, we replace the factor $16$ with $254$ and clip the count $n_{t+1}$ to $[-127,127]$.
\begin{equation}\label{eq:remainder}
\begin{aligned}
u_t &= w_t + \frac{n_t}{254}\,h(w_t) + \Delta w_t, \qquad
w_{t+1} = \mathrm{round}_{\mathrm{BF16}}(u_t), \\
n_{t+1} &= \operatorname{clip}_{[-127,127]}\!\left(\mathrm{sround}\!\left(\frac{254\,(u_t-w_{t+1})}{h(w_{t+1})}\right)\right).
\end{aligned}
\end{equation}

\subsection{Evaluation Protocol}

We evaluate our training checkpoints every 50 updates on the six mathematics test sets in Thinker's evaluation suite \citep{thinker}: MATH-500 \citep{math500}, AIME 2024, AIME 2025, AMC 2023, Minerva Math \citep{minerva}, and OlympiadBench \citep{olympiad}, comprising 1,547 test questions in total. At each checkpoint, every question receives one greedy response and eight independently seeded sampled responses (temperature 1, top-$p$ 1) generated with the Qwen-Math prompt at 1.5B and 7B and the native chat template at 27B, a 3,000-token response cap, and a 4,096-token context window. All responses are graded with the Dr.~GRPO boxed-answer grader \citep{drgrpo}, and the six-set mean is the unweighted average across the six mathematics test sets. We evaluate each released checkpoint with its authors' training prompt.

\subsection{Cost Estimation and Rental Prices}
\label{app:cost}

All monetary figures are reported in USD. Table~\ref{tab:rental} lists the hourly GPU rental rates used throughout the paper, computed as the equal-weighted mean of available quotes from Vast.ai, RunPod Community, AutoDL, and Lambda. To prevent a few unusually cheap listings from skewing the average, Vast.ai is represented by the median of its available verified offers. Quotes were collected on 25 November 2026, and undiscounted AutoDL rates were converted using the 24 November exchange rate of 6.6645 CNY per USD. Container-only and monthly-contract offers are excluded, as are separate evaluation, idle time, and storage charges.

\begin{table}[ht]\centering\small
\caption{\DualverseCaptionTitle{GPU rental prices used for cost estimation.} Prices are USD per GPU-hour. The final column is the equal-weight mean of the available provider quotes.}
\label{tab:rental}
\begin{DVRentalTable}
\DVHeaderRow \DVHead{l}{GPU} & \DVProviderHead{\shortstack{Vast.ai\\median}} & \DVProviderHead{\shortstack{RunPod\\Community}} & \DVProviderHead{AutoDL} & \DVProviderHead{Lambda} & \DVHead{r}{Mean} \\
\DVFirstRow{RTX 5090 32 GB & 0.760 & 0.690 & 0.486 & -- & 0.645} \\
H100 SXM 80 GB & 2.267 & 2.690 & -- & 4.890 & 3.282 \\
H100 PCIe 80 GB & 2.667 & 1.990 & -- & 3.790 & 2.816 \\
A100 SXM 80 GB & 1.031 & 1.390 & -- & 3.190 & 1.870 \\
A100 PCIe 80 GB & 0.867 & 1.190 & -- & -- & 1.028 \\
\DVLastRow{A100 PCIe 40 GB & -- & -- & 0.518 & -- & 0.518} \\
\end{DVRentalTable}
\end{table}

We match each run to its exact GPU variant whenever known. When external disclosures specify only H100 or A100 80\,GB without a form factor, we apply the corresponding SXM rate.

\paragraph{Cost evaluation of codebases.}
For each codebase in Table~\ref{tab:a1}, we find the smallest number of H100s or RTX~5090s (up to eight of each type) that can complete an RL update with 16 prompts and eight responses per prompt. We exclude one warm-up update and use the mean time of the next ten to estimate the cost of 100 updates.

\paragraph{Released models.}
We use author-disclosed dollar totals when available; otherwise, we multiply disclosed GPU counts and runtimes by the rates in Table~\ref{tab:rental}. For PRIME-Zero, we use the authors' \href{https://github.com/PRIME-RL/PRIME/issues/17\#issuecomment-2585210458}{three-day runtime for PRIME} as a proxy. When runtime is not reported, we profile the released repository and extrapolate to its training schedule.

\section{Additional Results}
\label{app:results}

\subsection{Memory and Time Profiles}
\label{app:profiles}

\begin{figure}[ht]
\centering
\fig{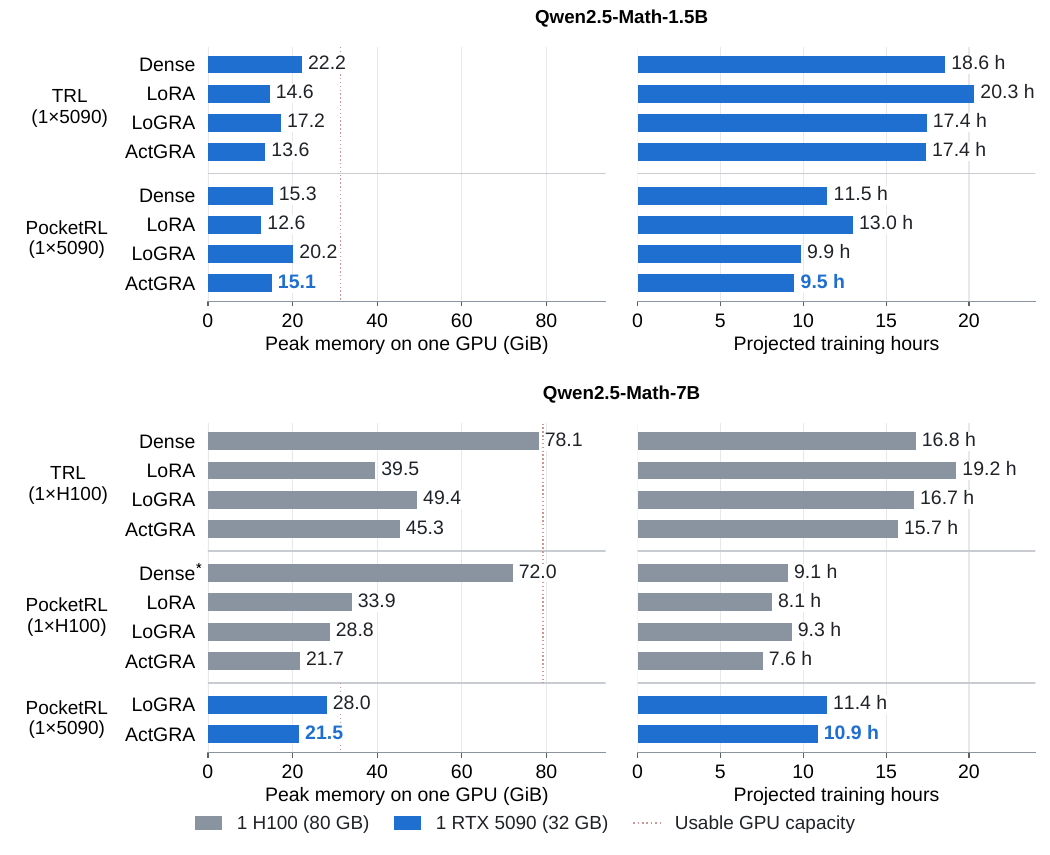}{fig4_memory_time.png}
\caption{\DualverseCaptionTitle{Peak GPU memory and projected training hours on a single GPU.} At 7B, LoRA runs out of memory on one RTX 5090 in both codebases. \textsuperscript{*}This profile uses BF16 optimizer moments for the weight matrices to fit on one H100.}
\label{fig:memtime}
\end{figure}

To disentangle the effect of codebase and gradient compression method, we implemented Dense, LoRA, LoGRA, and ActGRA in both TRL with vLLM~\citep{trl}\citep{vllm} and \sys{}. We profiled ten updates on a single GPU and linearly extrapolated the measured time to estimate the training hours for a complete run (Figure~\ref{fig:memtime}).

At 1.5B, \sys{} ActGRA on an RTX 5090 is the fastest configuration at 9.5~h, whereas TRL ActGRA on the same GPU takes 17.4~h (84\% longer). More detailed profiling shows that TRL generates answers faster, but \sys{} saves time overall in the remaining steps, which include training and the additional weight synchronization that TRL requires for inference.

At 7B, \sys{} ActGRA takes an estimated 7.6~h on one H100 SXM, compared with 8.1~h for \sys{} LoRA. Weight sharing also lowers memory usage, allowing 7B LoGRA and ActGRA in \sys{} to fit on a single RTX 5090 at 28.0 and 21.5~GiB, respectively. Switching to an RTX 5090 increases training time by 44\%, from 7.6 to 10.9~h. However, the lower price of the RTX 5090 cuts the estimated cost by 72\%, from \$24.79 to \$7.01, making it the cheapest 7B option.

\FloatBarrier
\section{Additional Analysis}

\subsection{Activation Concentration and Basis Refitting}

\begin{figure}[ht]
\centering
\includegraphics[width=\linewidth]{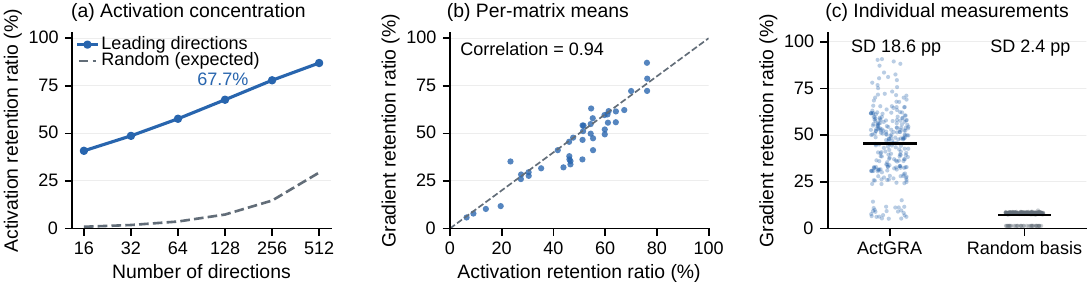}
\caption{\DualverseCaptionTitle{Activation concentration and gradient retention on Qwen2.5-Math-1.5B.} (a) Activation retention across ranks. (b) Activation and gradient retention at rank 128, averaged within each weight matrix. (c) Individual gradient retention ratios for ActGRA and a random basis at rank 128. Black bars mark the means.}
\label{fig:scatter}
\end{figure}

We analyze how concentrated the input activations are across the base model and two trained checkpoints. Figure~\ref{fig:scatter}a shows the activation retention ratio---the fraction of the squared activation norm captured along the leading directions of the activations. On average, the leading 128 directions already capture 67.7\% of the squared norm, whereas a random subspace of the same dimension is expected to capture only 7.3\%. This concentration supports the use of an activation-fitted projection in ActGRA.

We then investigate whether capturing the activations also captures the gradient. Because weight gradients are weighted sums of input activations, a subspace could in principle capture a large portion of the activations while missing the gradient if the upstream gradients give little weight to those directions. In practice, however, a basis that captures more of the activations generally also captures more of the gradient. Figure~\ref{fig:scatter}b shows a Pearson correlation of 0.94 between the activation and gradient retention ratios across 42 weight matrices, each averaged over six measurements. This relationship supports using activations alone to choose directions that also retain the gradient.

Finally, Figure~\ref{fig:scatter}c plots the 252 individual gradient retention ratio measurements per method for ActGRA and a random basis across weight matrices. ActGRA exhibits a larger standard deviation in gradient retention than the random basis (18.6 vs.\ 2.4 percentage points), plausibly because the concentration of activations varies across weight matrices. Nonetheless, ActGRA retains a much larger portion of the gradient on average than the random basis (45.5\% vs.\ 7.3\%).

\subsection{Gradient Retention Measurement}

In this section, we explain how we measure the gradient retention ratio in Figure~\ref{fig:gradshare}a and Figure~\ref{fig:scatter}b and~\ref{fig:scatter}c. We evaluate Qwen2.5-Math-1.5B at updates 0 (the base model), 50, and 250. For each checkpoint, we sample two batches of 16 prompts with eight responses per prompt, run a backward pass, and accumulate the full weight gradients in FP32. We measure the query, key, value, attention-output, gate, up, and down projections in transformer blocks 0, 4, 10, 14, 22, and 27, giving 42 weight matrices.

At each rank $r$, we fit the ActGRA basis $Q$ to one batch using the proposed method and measure retention on the other batch as $\|GQ\|_F^2 / \|G\|_F^2$, where $G$ is the full weight gradient, then do the same in reverse order. For random projection, we orthonormalize $r$ random sign vectors. At each rank, we average the individual retention ratios over 42 weight matrices, three checkpoints, and both batches.

\subsection{Hyperparameter Sensitivity}

We perform an additional hyperparameter sensitivity analysis on Qwen2.5-Math-1.5B, reporting MATH-500 avg@8 accuracy in Table~\ref{tab:hyperparameter-sensitivity}. First, decreasing the ActGRA rank from 128 to 32 leads to a minor performance drop from 68.2\% to 67.7\%. We also vary the learning rate. Increasing the default ($10^{-6}$) fourfold yields similar final performance (68.8\%), whereas increasing it further causes accuracy to deteriorate substantially (51.7\% at $10^{-5}$). Finally, using the predicted-KL controller proposed in LoGRA with ActGRA leads to worse performance at 65.9\%.

We also evaluated LoRA in our \sys{} codebase. At each method's best tested learning rate, LoRA performs similarly to ActGRA (68.8\% vs.\ 68.8\%, Table~\ref{tab:hyperparameter-sensitivity}). In the 7B \sys{} profile at matched rank~128, LoRA uses substantially more memory. LoGRA also reports higher peak training memory for LoRA than for LoGRA~\citep{logra}. LoRA runs out of memory on a single RTX~5090, while ActGRA uses 21.5~GiB (Figure~\ref{fig:memtime}). We therefore exclude LoRA from the main runs.

\begin{table}[ht]\centering\small
\caption{\DualverseCaptionTitle{Hyperparameter sensitivity on Qwen2.5-Math-1.5B.} MATH-500 avg@8 accuracy at update 500.}
\label{tab:hyperparameter-sensitivity}
\begin{DVSensitivityTable}
\DVHeaderRow \DVHead{l}{Method} & \DVHead{l}{Setting} & \DVHead{r}{MATH-500 avg@8 (\%)} \\
\DVFirstRow{ActGRA & Rank 128 (default) & 68.2} \\
ActGRA & Rank 32 & 67.7 \\
ActGRA & Learning rate $10^{-6}$ (default) & 68.2 \\
ActGRA & Learning rate $4\times10^{-6}$ & 68.8 \\
ActGRA & Learning rate $10^{-5}$ & 51.7 \\
ActGRA & Predicted-KL controller & 65.9 \\
\midrule
LoGRA & Rank 32 & 67.1 \\
LoGRA & Rank 256 & 66.6 \\
\midrule
LoRA & Learning rate $2\times10^{-5}$ & 68.8 \\
LoRA & Learning rate $5\times10^{-5}$ & 64.8 \\
\DVLastRow{LoRA & Learning rate $2\times10^{-4}$ & Crashed} \\
\end{DVSensitivityTable}
\end{table}
\FloatBarrier

\subsection{Answer Length}
\label{app:length}

\begin{figure}[ht]
\centering
\includegraphics[width=0.84\linewidth]{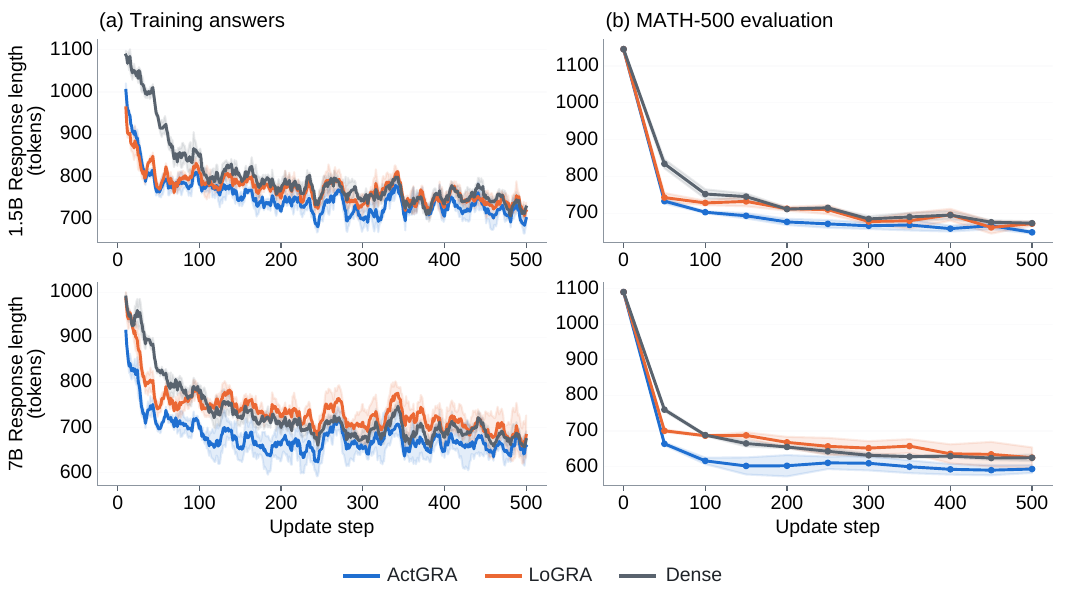}
\caption{\DualverseCaptionTitle{Response length during training and evaluation.} Mean response length in tokens for training rollouts (left) and MATH-500 evaluations (right). Lines show the mean across three seeds and shading shows one standard error.}\label{fig:length}
\end{figure}

Figure~\ref{fig:length} tracks the mean response length during training and at each evaluated checkpoint. ActGRA, LoGRA, and Dense training all shorten responses relative to the base models, which average roughly 1,100 tokens. ActGRA produces slightly shorter answers than LoGRA and Dense at convergence: at update 500 on MATH-500, mean answer lengths are 648, 672, and 673 tokens at 1.5B, and 593, 625, and 625 tokens at 7B, respectively.

\section{Limitations}

Our empirical evidence is limited to six mathematics test sets evaluated on one model family at the 1.5B, 7B, and 27B scales. Evaluating even larger models would be valuable, but it was beyond our available resources. The cost savings also depend on GPU rental prices and availability, both of which fluctuate and are outside our control.

\end{document}